\documentclass[fleqn,10pt]{wlpeerj}

\usepackage{url}
\usepackage{subfig}

\usepackage{color}

\title{Check-worthy Claim Detection across Topics for Automated Fact-checking}

\author[1,2]{Amani S. Abumansour}
\author[1]{Arkaitz Zubiaga}
\affil[1]{Queen Mary University of London, United Kingdom}
\affil[2]{Taif University, Saudi Arabia}

\begin{abstract}
An important component of an automated fact-checking system is the claim check-worthiness detection system, which ranks sentences by prioritising them based on their need to be checked. Despite a body of research tackling the task, previous research has overlooked the challenging nature of identifying check-worthy claims across different topics. In this paper, we assess and quantify the challenge of detecting check-worthy claims for new, unseen topics. After highlighting the problem, we propose the AraCWA model to mitigate the performance deterioration when detecting check-worthy claims across topics. The AraCWA model enables boosting the performance for new topics by incorporating two components for few-shot learning and data augmentation. Using a publicly available dataset of Arabic tweets consisting of 14 different topics, we demonstrate that our proposed data augmentation strategy achieves substantial improvements across topics overall, where the extent of the improvement varies across topics. Further, we analyse the semantic similarities between topics, suggesting that the similarity metric could be used as a proxy to determine the difficulty level of an unseen topic prior to undertaking the task of labelling the underlying sentences.
\end{abstract}

\begin{document}

\flushbottom
\maketitle
\thispagestyle{empty}

\section*{Introduction}

Where the availability of online information continues to grow, assessing the accuracy of this information can be a challenge, where support from automated fact-checking systems is becoming increasingly necessary to help getting rid of misinformation \citep{zeng2021automated}. One of the key components of an automated fact-checking pipeline is the check-worthy claim detection system, which given a collection of sentences as input, ranks them based on their need to be fact-checked \citep{clef-checkthat:2021:task1}. The ranked output can be used in different ways, among others by providing fact-checkers with the most important claims to work on, by feeding a claim verification system with the top claims to be checked, or by informing the general public that the top claims may be disputed and need verification.

In recent years, there has been a body of research developing models for check-worthy claim detection \citep{konstantinovskiy2021toward}. Many of these systems have been tested against each other by competing on benchmark datasets released as part of the CheckThat! shared tasks \citep{clef-checkthat:2021:task1}. The recent tendency of these models has been to use Transformer models such as BERT and RoBERTa which have led to improved performance for detecting check-worthy claims in different languages including English and Arabic \citep{nakov2022overview}. Existing claim detection models have been developed for and tested in datasets that provide a mix of topics in both training and test data, in such a way that topics encountered in the test set were already seen during training. While this has provided a valuable setting to compare models against each other, it is also arguably far from a realistic scenario where a claim detection model is trained from data pertaining to certain topics to then detect check-worthy claims in a new, unseen topic; for example, where a model trained from claims associated with politics needs to be used for detecting claims associated with COVID-19 when the new topic emerges.

The objective of our work is to assess the extent of the problem of detecting check-worthy claims across topics, i.e. simulating the scenario where new topics emerge and for which labelled data is lacking or limited. To undertake this challenge, we use an Arabic language dataset from the CheckThat! shared tasks which comprises a collection of claims and non-claims distributed across 14 different topics. We first conduct experiments in zero-shot settings using a model called AraCW, which has not seen any data associated with the target topic; the low performance of this model for some of the topics demonstrates the importance of this hitherto unstudied problem in check-worthy claim detection. We further propose AraCWA, an improved model that incorporates two additional components to enable few-shot learning and data augmentation. We test AraCWA with a range of few-shot and data augmentation settings to evaluate its effectiveness, achieving improvements of up to 11\% over AraCW.

To the best of our knowledge, our work is the first to tackle the check-worthy claim detection task across topics. In doing so, we make the following novel contributions:

\begin{itemize}
 \item We establish a methodology for cross-topic claim detection in both zero-shot and few-shot settings.
 \item We assess the extent of the challenge of detecting claims for new, unseen topics through zero-shot experiments.
 \item We propose an improved approach, AraCWA, which leverages few-shot learning in combination with data augmentation strategies to boost the performance on new topics.
 \item We perform a set of ablative experiments to assess the contribution of the data augmentation technique as well as the few-shot strategy.
 \item We perform an analysis of the characteristics of the topics under study, finding that the semantic similarity between topics can be a reasonable proxy to determine the difficulty level of a topic.
\end{itemize}

In presenting the first study in cross-topic claim detection, our work also posits the need for more research in this challenging yet realistic scenario.

\section*{Related Work on Check-Worthy Claim Detection}

While a large portion of the research in automated fact-checking has focused on claim verification \citep{pradeep2021scientific,schuster2021get,zeng2022active}, work on check-worthy claim detection has been more limited. However, the claim detection task constitutes an important first step that identifies the claims to be prioritised and to be fed to the subsequent verification component \citep{hassan2017toward}. In what follows, we discuss existing work in check-worthy claim detection with a particular focus on the Arabic language and data augmentation.

Existing research in check-worthy claim detection has primarily focused on sentences in the English language \citep{zeng2021automated}. This is the case, for example, of one of the first fact-checking systems developed in the field, ClaimBuster \citep{hassan2017claimbuster,hassan2017toward}, which studied detection of claims from transcripts of US presidential election debates through what they called ``claim spotter''. In another study, \citet{atanasova2019automatic} investigated the use of context and discourse features to improve the claim detection component; while this is a clever approach, it does not generalise to other types of datasets such as tweets in our case, given the limited context available typically in social media. More recently, \citet{konstantinovskiy2021toward} developed a model, namely CNC, that leverages InferSent embeddings \citep{conneau2017supervised} along with part-of-speech tags and named entities to identify claims.

In addition to the above studies, there have been numerous competitive efforts to develop claim detection systems thanks to the CheckThat! series of shared tasks \citep{elsayed2019overview,clef-checkthat:2021:task1}. These shared tasks have provided a benchmark for researchers to study the task in a competitive manner, including publicly available datasets in multiple languages. In our study, we use the datasets available from these shared tasks in the Arabic language.

\paragraph{Check-worthy claim detection in the Arabic language.}  Check-worthy claim detection for the Arabic language was first tackled by ClaimRank \citep{jaradat-etal-2018-claimrank} when training the neural network model with originally English sentences from political debates translated into Arabic. In addition, other learning models, e.g.  Gradient boosting and k-nearest neighbors, were inspected for the task in Arabic of check-worthiness at CheckThat! 2018 \citep{atanasova2018overview}.

The paradigm shift in the field of NLP occurred with the emergence of Transformers \citep{wolf2020transformers}, and Bidirectional Encoder Representations (BERT), which became the state-of-art model for several tasks, including a range of text classification tasks \citep{devlin-etal-2019-bert}. This is also reflected in the works on check-worthy claim detection as most participants of the recent CheckThat Labs fine-tuned BERT models for Arabic such as mBERT, AraBERT, and  BERT-Base-Arabic \citep{Hasanain2020OverviewOC,clef-checkthat:2021:task1,nakov2022overview}.

However, existing models focus on evaluating their performance on test sets whose topics overlap with those seen in the training data, without focusing on the challenges that different topics may pose to the task. As part of this study, we focus on the novel challenge of investigating how claim detection models perform differently for different topics.

\paragraph{\textbf{Data augmentation for claim detection.}} Data augmentation techniques enable to leverage the (generally limited) existing data to generate new synthetic data through alterations. By carefully altering the original data samples, one can augment the available training data with the newly generated samples; however, data augmentation techniques may also produce noisy samples which lead to performance deterioration, and therefore creation of useful samples is crucial. Data augmentation strategies have also been successfully used to alleviate different problems, e.g. to settle an imbalanced dataset and to reduce model biases \citep{feng2021survey}. Different data augmentation methods exist, which \citet{li2022data} categorised into three types: (i) paraphrasing, where the new augmented sentence has similar semantics to the original one, (ii) noising, which adds discrete or continuous noise to the sentence, and (iii) sampling, which is similar to paraphrasing but for more specific tasks which need more information about this to this task like data format.

Data augmentation techniques have been barely used in claim detection. For example, \citet{williams2021accenture} used a contextual word embedding augmentation strategy in their participation in the 2021 CheckThat! lab. These data augmentation techniques have however not been studied studied in the context of cross-topic claim detection. In our study, we propose the AraCWA model which provides the flexibility of being used with different data augmentation strategies. More specifically, we test three data augmentation strategies to evaluate their effectiveness in the cross-topic claim detection task, namely back-translation, contextual word embedding augmentation and text-generating augmentation.

\section*{Methodology}
\label{sec:Methodology}

\subsection*{Problem Formulation and Evaluation}
\label{sec:Problem formulation}

The check-worthy claim detection task consists in determining which of the sentences, out of a collection, constitute claims that should be fact-checked. Given a collection of sentences $S = \{s_1, s_2, ..., s_n\}$, the model needs to classify each of them into one of $L = \{CW, NCW\}$, where CW = check-worthy and NCW = not check-worthy. The collection of sentences in $S$ belongs to a number of topics $T = \{t_1, t_2, ..., t_m\}$, where each sentence $s_j$ only belongs to one topic $t_i$. Where previous research has experimented with a mix of sentences pertaining to all topics spread across both training and test data, our objective here to experiment with claim detection across topic is to leave a topic $t_i$ out for the test, using the sentences from the rest of the topics for training.

While the task has sometimes been formulated as a classification problem, here and in line with the CheckThat! shared tasks, we formulate it as a ranking task. As a ranking problem, the model needs to produce an ordered list of sentences in $S$, where the sentences in the top of the list are predicted as the most likely to be check-worthy claims. The task is in turn evaluated through the Mean Average Precision (MAP) metric, which is defined as follows:

\begin{equation}
    MAP = \frac{1}{k} \sum_{i=1}^k AP_i
\end{equation}

That is the average of the AP's (average precisions) across classes, where in the case of claim detection $k = 2$.

AP is in turn defined as follows:

\begin{equation}
    AP_c = \frac{1}{N} \sum{j=1}^N P_i
\end{equation}

where N is the number of instances being considered in the evaluation, whose precision values are averaged.

Ultimately, $AP_c$ measures the average precision on the top $N$ items for class $c$. After calculating the AP values for both classes, they are averaged to calculate the final MAP.

\subsection*{Dataset}
\label{sec:dataset}

We make combined use of two claim detection datasets in the Arabic language, CT20-AR and CT21-AR, released as part of two different editions of the CheckThat! shared task. The datasets contain tweets which are labelled as check-worthy or not check-worthy. The tweets are in turn categorised into topics, which allows us to set up the cross-topic experiments. We show the main statistics of these two datasets in Table \ref{tab:AR-datasets}. 

\paragraph{\textbf{The CT20-AR dataset.}} This was released as part of the CheckThat! 2020 Lab \citep{Hasanain2020OverviewOC} and it comprises 7.5k tweets distributed across 15 topics, with the tweets evenly distributed across the topics, i.e. 500 tweets per topic.

\paragraph{\textbf{The CT21-AR dataset.}} Released as part of the CheckThat! 2021 Lab \citep{nakov2021clef}, part of this dataset overlaps with CT20-AR. After removing the overlapping tweets, this dataset provides two new topics with a total of 600 tweets (347 and 253 tweets for each topic).

The combination of both datasets originally led to 17 topics (15 + 2). However, during the process of combining both datasets into one, we found that four of the topics are related to COVID-19, namely those with topic IDs ``CT20-AR-03'', ``CT20-AR-28 w1'', ``CT20-AR-28 w2'' and ``CT20-AR-29'', Given the significant overlap between them, we combined these four topics into one, hence reducing the number of topics from 17 to 14. The final, combined dataset contains 8,100 tweets distributed across 14 topics. See Appendix A for a list and description of these topics.

The CT20-AR and CT21-AR datasets differ slightly in the labels. Where the former only provides binary labels for check-worthy (CW) and not check-worthy (NCW), the latter provides two levels of labels: (i) claim (C) or not claim (NC), and (ii) check-worthy (CW) and not check-worthy (NCW). In both cases, we only rely on the CW / NCW labels for consistency.

\begin{table}[ht]
 \centering
 \begin{tabular}{l|c|c|c|c}
    Dataset & Year & Num. of tweets & Num. of topics &Labels \\\hline
        CT20-AR \citep{Hasanain2020OverviewOC}
            &2020
                &7,500 
                    & 15 
                        &CW / NCW\\
        CT21-AR \citep{clef-checkthat:2021:task1}
            &2021
                & 600 
                    & 2 
                        &C / NC, CW / NCW\\

 \end{tabular}
 \caption{Arabic Datasets for claim check-worthiness detection. C: claim, NC: non-claim, CW: check-worthy, NCW: non-check-worthy}
 \label{tab:AR-datasets}
\end{table}

\begin{figure}[ht]
\centering
\includegraphics[
  width=0.5\textwidth
]{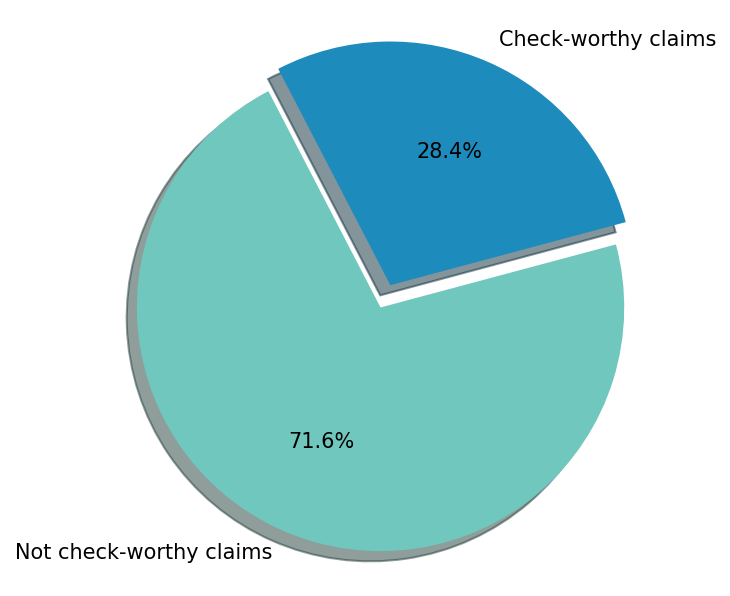}
\caption{Overall proportion of check-worthy and not check-worthy claims for CT20-AR and CT21-AR datasets.}
\label{fig:OverallDatasets}
\end{figure}

The distribution of labels in the entire dataset is clearly skewed towards the NCW class accounting for 71.6\% of the tweets, whereas the CW class accounts for 28.4\% of the tweets (see Figure \ref{fig:OverallDatasets}).

\begin{figure}[ht]
\centering
\includegraphics[
width=15.5cm, height=9cm
]{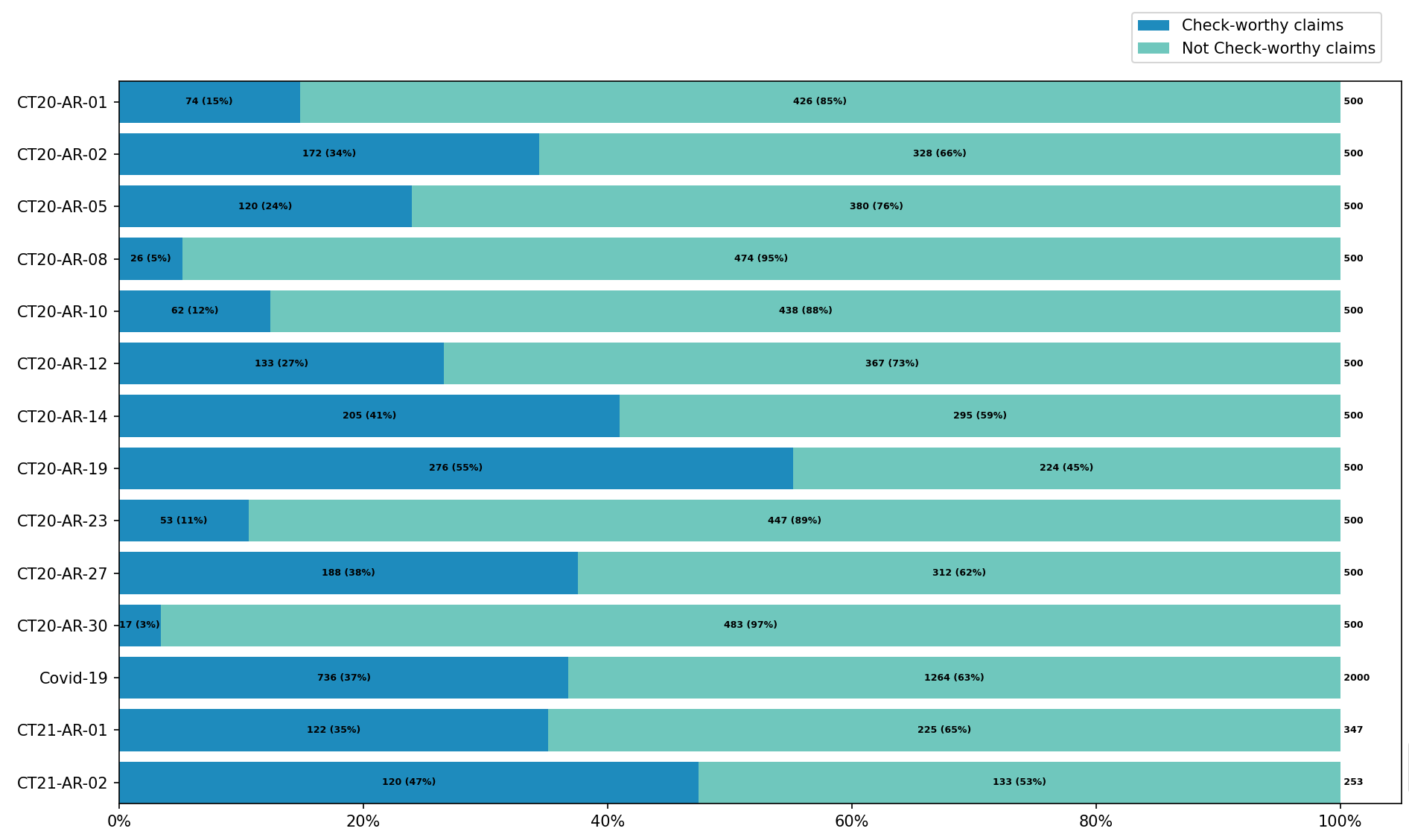} 
\caption{Distribution of check-worthy and not check-worthy claims over topics of CT20-AR and CT21-AR datasets.}
\label{fig:ClassesofTopics}
\end{figure}

Further drilling down into the distribution of CW and NCW across the different topics, Figure \ref{fig:ClassesofTopics} shows a high degree of variation. With only one topic where the CW class is larger than the NCW class (i.e. CT20-AR-19, ``Turkey's intervention in Syria''), all of the other classes have different degrees of prevalence of the NCW class. Some of the topics have a particularly low number of CW instances, which is the case of ``CT20-AR-08" (i.e. ``Feminists") with only 5\% of check-worthy claims, and ``CT20-AR-30" (i.e. ``Boycotting countries and spreading rumors against Qatar") with only 3\% of check-worthy claims.

\subsection*{Experiment settings}
\label{sec:Experiment settings}

We perform experiments in two different settings: (i) first, in zero-shot detection settings using AraCW, where the target topic is unseen during training, and (ii) then, in few-shot training settings using AraCWA, where a few samples of the target topic are seen during training.

To enable a fair comparison of experiments in both settings, we first hold out the data to be used as test sets. From each of the 14 topics in the dataset, we hold out 200 instances which will be used as part of the training data in the few-shot settings. Holding out these instances for all the experiments enables us to have identical test sets for both settings, where the 200 held-out instances are not used.

\paragraph{\textbf{Zero-shot detection across topics.}} In this case, we run a separate experiment for each topic. Where $t_i$ is the target topic in the test set, we train the model from all topics but $t_i$. The training data does not see any instances pertaining to topic $t_i$.

\paragraph{\textbf{Few-shot detection across topics.}} The setting in this case is similar to the zero-shot detection, where separate experiments are run for each topic $t_i$. However, the 200 instances held out above are used as part of the training data. Therefore, in this case, the experiments with topic $t_i$ in the test set use a model trained for all the other topics as well as 200 instances pertaining to the target topic $t_i$.

\subsection*{Claim Detection Models: AraCW and AraCWA}
\label{sec:Approaches}

We next describe the two different models we propose for our experiments. We propose AraCW as the base model for the zero-shot experiments, as well as the variant AraCWA which incorporates a data augmentation strategy on top of few-shot learning to boost performance across topics.

\paragraph{\textbf{AraCW.}} As one of the top-ranked models in the CheckThat! 2021 Lab, the AraCW model \citep{abumansour2021qmul} is built on top of a fine-tuned AraBERTv0.2-base \citep{antoun2020arabert} transformer model. We perform a set of preprocessing steps which proved effective during initial experiments through our participation in the CheckThat! 2021 Lab:

\begin{itemize}
 \item We substitute all URLs, email addresses, and user mentions with [url], [email] and [user] tokens.
 \item We eliminate line breaks and markup written in HTML, repeated characters, extra spaces, and unwanted characters including emoticons.  
 \item We correct white spaces between words and digits (non-Arabic, or English), and/or a combination of both, and before and after two brackets.
\end{itemize}

AraCW processes the tweets through Sequence Classification. Then, the results of the neural network output layer are passed into a softmax function in order to acquire the probability distribution for each predicted output class; we use the value output by the softmax function to rank the sentences by check-worthiness, which produces the final ranked output by ordering them by score.

\paragraph{\textbf{AraCWA.}} Our proposed AraCWA model builds on top of AraCW to incorporate two key components: few-shot learning and data augmentation. AraCWA aims to make the most of a small number of instances pertaining to the target topic by using a data augmentation strategy to increase its potential. With AraCWA, we also aim to assess the extent to which a small portion of labelled data can boost performance on new topics, where despite the cost of labelling data, it could be affordable given that it is a small portion.

AraCWA's additional components for few-shot learning and data augmentation are as follows:

\paragraph{\textit{Few-shot learning.}} This component leverages the 200 instances pertaining to the target topic, which were originally held out from the test set (see Experiment Settings). This enables to fine-tune the model in a broader training set which includes more relevant samples. The 200 instances incorporated into the training set add to the other 13 topics used for the training (i.e. excluding the 14th topic used as the target).

\paragraph{\textit{Data augmentation.}} Once the 200 instances are incorporated through the few-shot learning component, AraCWA leverages a data augmentation strategy to increase its presence by generating new synthetic samples. The data augmentation component of AraCWA is flexible in that it can make use of different data augmentation algorithms. In our experiments, we test AraCWA with three data augmentation methods, as follows:

\begin{enumerate}
     \item \textbf{Back translation:}\\
     The objective of a back translation strategy is to translate a text twice, first to another language, and then back to the original language. The new, back-translated sample is likely to be different while similar. To achieve this, we use the state-of-art OPUS-MT models for open translation      \citep{tiedemann-thottingal-2020-opus,tiedemann-2020-tatoeba}. We use these models to translate the sample data of a target topic into the English language, then translate them back to the Arabic language.
     These new, back-translated samples represent new training samples that augment the original training data.
     
     \item \textbf{Contextual word level augmentation:}\\
     The objective of this approach is to generate new samples by altering existing samples using a contextualised language model. In this case, we make use of the nlpaug \citep{ma2019nlpaug} python package to implement the contextual word embedding technique of augmentation on samples of a target topic. We make use of the arabertv02 transformer model to determine the best appropriate word for augmentation and set the proportion of words to be augmented to 0.3. Additionally, the parameter action is set to ``substitute," which will replace the word based on the results of the contextual embedding calculation.
     
     \item \textbf{Text generation:}\\
     The objective of this approach is to train a model that can then generate new texts from scratch. To achieve this, we use the aragpt2-medium \citep{antoun-etal-2021-aragpt2} model to implement AraGPT2, an Arabic version of the GTP2 model. 
     First, the samples from the target topics that we intend to generate more text from are prepared by ArabertPreprocessor for aragpt2-medium. We use the following settings with AraGPT2:
        \begin{itemize}
            \item num\_beam is set to 5 to execute beam search for cleaner text.
            \item maximum length of the generated text is 200.
            \item top\_p is set to 0.75 to determine the cumulative probability of most possible tokens to be selected for sampling.
            \item repetition\_penalty is set to 3 to penalize repetition in a text and avoid infinite loops.
            \item token length of no\_repeat\_ngram\_size is set to 3 in order to avoid repeating phrases.
        \end{itemize}       
\end{enumerate}

\section*{Results}\label{Results}

Next, we present results for our check-worthy claim detection experiments across topics, beginning with zero-shot experiments and followed by few-shot experiments, as well as ablation experiments where the data augmentation is removed from the pipeline.

\subsection*{AraCW: Zero-Shot detection across topics}
\label{sec:Results_zero-shot}

\begin{table}[htb]
 \centering
 \begin{tabular}{l|c|c|c|c}
  TopicID & Precision & Recall & F1 & MAP \\
  \hline
  CT20-AR-01 & 0.68 & 0.64 & 0.66 & 0.6408 \\
  CT20-AR-02 & 0.70 & 0.74 & 0.72 & 0.6473 \\
  CT20-AR-05 & 0.70 & 0.61 & 0.65 & 0.5983 \\
  CT20-AR-08 & 0.12 & 0.31 & 0.17 & {\color{red}\underline{0.2468}} \\
  CT20-AR-10 & 0.45 & 0.46 & 0.45 & {\color{red}\underline{0.3999}} \\
  CT20-AR-12 & 0.77 & 0.38 & 0.50 & 0.5637 \\
  CT20-AR-14 & 0.63 & 0.78 & 0.70 & 0.6563 \\
  CT20-AR-19 & 0.76 & 0.79 & 0.77 & \textbf{0.7538} \\
  CT20-AR-23 & 0.32 & 0.34 & 0.33 & {\color{red}\underline{0.2644}} \\
  CT20-AR-27 & 0.61 & 0.68 & 0.64 & 0.5647 \\
  CT20-AR-30 & 0.24 & 1.00 & 0.39 & {\color{red}\underline{0.4172}} \\
  Covid-19   & 0.71 & 0.74 & 0.72 & 0.6826 \\
  CT21-AR-01 & 0.73 & 0.42 & 0.54 & {\color{red}\underline{0.5429}} \\
  CT21-AR-02 & 0.56 & 0.92 & 0.70 & 0.6708 \\
  \hline
  Average    & 0.57 & 0.63 & 0.57 & 0.5464 \\
\end{tabular}
\caption{Results of our model's performance for TopicID as the target topic, where MAP is the main evaluation measure. The best result is shown in bold, while the performances below average are underlined and coloured in red.}
\label{tab:one topic-results}
\end{table}

Table \ref{tab:one topic-results} shows results for the zero-shot experiments using the AraCW model for the 14 topics separately. We can observe high variation in terms of MAP scores for the different topics, with values ranging from as low as 0.24 to 0.75. Where the average performance for all topics is over 0.54, we observe that five of the topics perform below this average (underlined and coloured in red in the table).

The best score is 0.7538 for the topicID ``CT20-AR-19” which is about ``Turkey's intervention in Syria” while the worst one is 0.2468 for ``CT20-AR-08” which includes tweets about ``Feminists”. In the first case, we assumed that the best performance is because of the nature of the political war on the topic, and there are some topics of the same nature that were used through training such as ``Houthis in Yemen" and ``Events in Libya". Thus, the model already learned some implications about the topic prior to testing.  In the second case, the topic of ``Feminists” is totally different and unique with respect to the other topics in the datasets most of which are political. Likewise, the model performed poorly with topicIDs ``CT20-AR-10” ``CT20-AR-23” ``CT20-AR-30” where the topics are about: ``Waseem Youssef” which refer to religious views, ``The case of the Bidoon in Kuwait” indicates a social issue, and ``Boycotting countries and spreading rumours against Qatar” contains rumours. The other results fluctuate between 54\% and 68\%.

This indicates that the AraCW model struggles in the zero-shot settings, particularly when the target topic differs substantially from the ones used for training. This in turn motivated the development of AraCWA leveraging few-shot learning and data augmentation, whose results we discuss in the next section.

\subsection*{Few-shot detection with data augmentation} 
\label{sec:Results_few-shot}

\begin{table}[ht]
 \centering
 \begin{tabular}{l|c|cc|cc|cc}
    \multicolumn{1}{c}{Topic ID} & \multicolumn{1}{|c}{AraCW} & \multicolumn{2}{|c}{AraCWA (BT)} & \multicolumn{2}{|c}{AraCWA (CWE)} & \multicolumn{2}{|c}{AraCWA (TxtGen)} \\
    \hline

    CT20-AR-01  & 0.6408  & 0.6664 & (+3\%) & 0.659  & (+2\%) & 0.6896  & (+5\%) \\
    CT20-AR-02  & 0.6473  & 0.7153 & (+7\%) & 0.7305 & (+8\%)   & 0.7245  & (+8\%)   \\
    CT20-AR-05  & 0.5983  & 0.5992 & (\underline{0\%})  & 0.5865  & (\underline{-1\%}) & 0.5845 & (\underline{-1\%})  \\
    CT20-AR-08  & {\color[HTML]{9A0000} 0.2468}  & 0.3707  & (+12\%)  & 0.4868  & ({\color[HTML]{3531FF}+24\%})  & 0.3751  & (+13\%)   \\
    CT20-AR-10  & {\color[HTML]{9A0000} 0.3999}  & 0.5288  & (+13\%) & 0.4252 & (+3\%)   & 0.4406 & (+4\%) \\
    CT20-AR-12 & 0.5637  & 0.8345  & (+27\%) & 0.8448  & ({\color[HTML]{3531FF}+28\%})  & 0.8623     & (+30\%)    \\
    CT20-AR-14  & 0.6563  & 0.7342 & (+8\%)   & 0.7729  & (+12\%)  & 0.7264  & (+7\%) \\
    CT20-AR-19  & 0.7538   & 0.8444 & (+9\%)  & 0.863  & (+11\%)  & 0.8611  & (+11\%)  \\
    CT20-AR-23  & {\color[HTML]{9A0000} 0.2644} & 0.3063  & (+4\%)   & 0.2616  & (\underline{0\%})  & 0.2628 & (\underline{0\%})  \\
    CT20-AR-27 & 0.5647 & 0.6248  & (+6\%)  & 0.6222 & (+6\%)  & 0.6077 & (+4\%)   \\
    CT20-AR-30  & {\color[HTML]{9A0000} 0.4172} & 0.6091  & (+19\%)  & 0.6264  & ({\color[HTML]{3531FF}+21\%}) & 0.5797 & (+16\%)  \\
    Covid-19 & 0.6826  & 0.7151 & (+3\%) & 0.6988  & (+2\%)  & 0.7001  & (+2\%) \\
    CT21-AR-01 & 0.5429 & 0.7382 & (+20\%) & 0.7884 & ({\color[HTML]{3531FF}+25\%}) & 0.7814 & (+24\%) \\
    CT21-AR-02  & 0.6708 & 0.7865 & (+12\%)  & 0.8438  & (+17\%) & 0.7721  & (+10\%) \\
    \hline
    Average & 0.5464 &0.6481 & (+10\%) & 0.6579 & (+11\%) & 0.6406 & (+9\%) \\
\end{tabular}
\caption{Improvement over topics for each technique on 200 samples. (FSL) denotes Few-shot learning, then integrated with data augmentation techniques that are Back-translation(BT), Contextual word Embedding(CWE), and text generation (TxtGen). The topics that scored poor performance in the zero-shot experiment are shown in Red. The high percentage of improvements in FSL+CWE is displayed in blue. Underlined percentages indicate zero or below enhancement.}
\label{tab:results-across-topics}
\end{table}

Table \ref{tab:results-across-topics} shows results for the AraCWA model leveraging few-shot learning and data augmentation, namely Back-translation (BT), Contextual word Embedding (CWE), and text generation (TxtGen). We observe substantial improvements of the three AraCWA variants over the AraCW baseline model, with overall improvements ranging from 9\% with AraCWA/TxtGen to 11\% with AraCWA/CWE. These results demonstrate that AraCWA's enhanced ability to incorporate few shots and augment data leads to substantial improvements.

When we look at the improvements across the individual topics (indicated as percentages within brackets reflecting the absolute improvement over AraCW), we observe that most of the differences are positive. Exceptions include the topicID ``CT20-AR-05" where performance drops by 1\% and the topicID ``CT20-AR-23" about ``Bidoon in Kuwait", where the performance difference is negligible.

Apart from these exceptions, the rest of the topics experience a positive impact with the use of AraCWA. Most remarkably, four of the topics experience an improvement above 20\%, which are coloured in blue. Looking back at the topics that proved problematic with the use of the AraCW model (highlighted in red in the AraCW column), we observe that the topicID ``CT20-AR-10" (``Waseem Youssef")  improves by 3\% with AraCWA/CWE, the topicID ``CT20-AR-08" (``Feminists") improves by 24\% and the topicID ``CT20-AR-30" (``boycotting countries and spreading rumors against Qatar") improves by 21\%. The remainder topic, ``CT20-AR-23" (``Bidoon in Kuwait") has no improvement (0\%). Other topics with remarkable improvements include ``CT21-AR-01" (``Events in Gulf; Saudi-Qatari reconciliation and Gulf summit") which improves by 25\%, and the topicID ``CT20-AR-12" (``Sudan and normalization") with han improvement of 28\%. All in all, the overall 11\% improvement of AraCWA/CWE shows the great potential of AraCWA compared to AraCW, which almost certainly guarantess that performance will not drop, but its improvement margin varies substantially.

\subsection*{Ablation experiments: Few-shot learning without data augmentation}

\begin{table}[htb]
 \centering
 \begin{tabular}{l|c|cc}
    & \multicolumn{1}{|c}{AraCWA} & \multicolumn{2}{|c}{} \\
    \multicolumn{1}{c}{Topic ID} & \multicolumn{1}{|c}{(no DA)} & \multicolumn{2}{|c}{AraCWA (CWE)} \\
    \hline
    CT20-AR-01 & 0.6883 & 0.6590 & (-3\%) \\
    CT20-AR-02 & 0.6935 & 0.7305 & (+4\%) \\
    CT20-AR-05 & 0.6002 & 0.5865 & (-1\%) \\
    CT20-AR-08 & 0.3796 & 0.4868 & (+11\%) \\
    CT20-AR-10 & 0.4660 & 0.4252 & (-4\%) \\
    CT20-AR-12 & 0.8467 & 0.8448 & (0\%) \\
    CT20-AR-14 & 0.7354 & 0.7729 & (+4\%) \\
    CT20-AR-19 & 0.8497 & 0.8630 & (+1\%) \\
    CT20-AR-23 & 0.3723 & 0.2616 & (-11\%) \\
    CT20-AR-27 & 0.6403 & 0.6222 & (-2\%) \\
    CT20-AR-30 & 0.5730 & 0.6264 & (+5\%) \\
    Covid-19   & 0.7101 & 0.6988 & (-2\%) \\
    CT21-AR-01 & 0.6471 & 0.7884 & (+14\%) \\
    CT21-AR-02 & 0.8554 & 0.8438 & (-1\%) \\   
    \hline
    Average    & 0.6469 & 0.6579 & (+1\%) \\
 \end{tabular}
 \caption{Performance scores for AraCWA with the CWE data augmentation strategy, and without any data augmentation (no DA).}
 \label{tab:ablation}
\end{table}

As part of the first ablation experiment to test the different components of AraCWA, we test it with the data augmentation component. Table \ref{tab:ablation} shows the results comparing the best version of AraCWA using the CWE data augmentation with the ablated AraCWA which does not use any data augmentation strategy.

We observe that the use of the data augmentation strategy leads to an overall improvement of 1\% over the non-augmented baseline. There are three topics where the data augmentation makes a major impact, with improvements of 11\% and 14\% for CT20-AR-08 and CT21-AR-01 respectively, and a drop of 11\% for CT20-AR-23. The rest of the topics experience a lesser impact, with changes ranging from -4\% to +5\%.

\subsection*{Ablation experiments: Using fewer shots}

\begin{figure}[htb]
\centering
\includegraphics[width=0.75\linewidth
]
{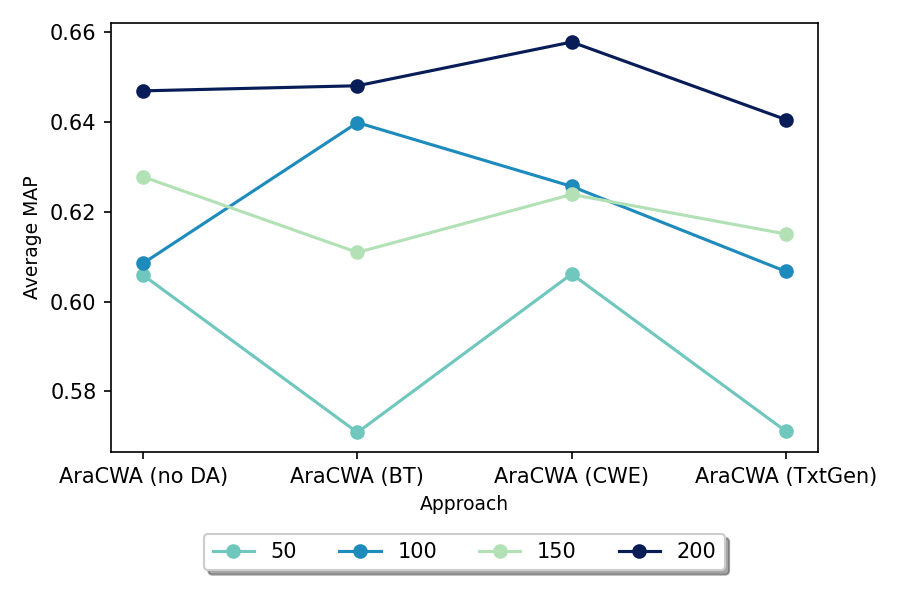} 
\caption{Performance scores for AraCWA with different numbers of shots ranging from 50 to 200.}
\label{fig:fewer-shots}
\end{figure}

We next test another ablated version of AraCWA, where we reduce the number of shots fed to the few-shot component from the original 200 shots. In this case, we test with 50, 100, 150 and 200 shots. Figure \ref{fig:fewer-shots} shows the results for AraCWA with different numbers of shots. While there is some inconsistency between the models using 100 and 150 shots, we see that the model with 50 shots performs clearly the worst, whereas the model with 200 shots performs the best. These experiments show the importance of the few-shot learning component of AraCWA, which contributes the most to the improvement of AraCWA over AraCW.

\section*{Discussion}
\label{sec:Discussion}

Despite the overall improvement of AraCWA, our experiments have shown substantial variability in performance across different topics in the claim detection task, suggesting that indeed some topics are much more challenging than others. Despite the clear improvement of AraCWA over AraCW, the improved model still struggles with some of the topics. One of the factors determining how easy or difficult a new topic will be is its similarity or dissimilarity with respect to the topics seen by the model.

Motivated by this, we next look at the semantic similarities between topics and we analyse how they may help us explain the trends in performance we observe. We use the AraBERT transformer model to compute pairwise semantic similarities between the 14 topics in the dataset, which helps us quantify how topics resemble / differ between them.

\begin{figure}[ht]
\centering
\includegraphics[
  width=\linewidth
]{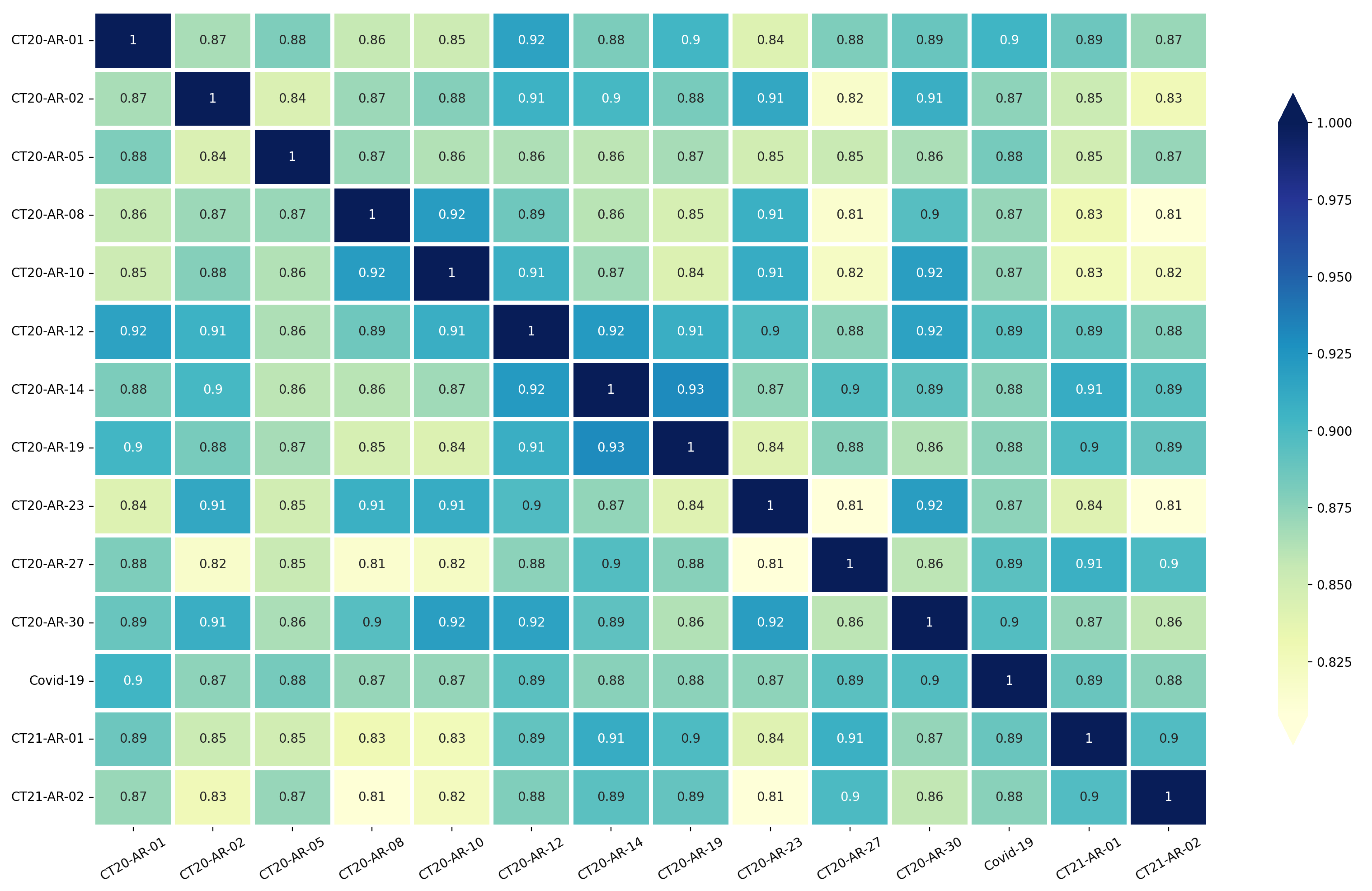}
\caption{Pairwise similarities between the 14 topics in our dataset.}
\label{fig:Topics_Similarities}
\end{figure}

Figure \ref{fig:Topics_Similarities} shows the pairwise semantic similarity scores for the 14 topics. While the similarity scores range from low 0.80's to mid 0.90's, we observe that the lower values are clustered around particular topics. This is for example the case for topics like ``CT20-AR-08" (``Feminists''), and ``CT20-AR-23" (``The case of the Bidoon in Kuwait"), both of which have a large number of low similarity scores and also showed lower performance scores in the zero-shot learning experiments with AraCW as shown in Table \ref{tab:one topic-results}.

Even if some of the above similarity scores may explain the performance observed in our experiments, there are some exceptions. For example, topicID ``CT20-AR-30" (``Boycotting countries and spreading rumors against Qatar") obtains consistently high similarity scores above 86\% while it proved to be a challenging topic in the experiments. While the semantic similarity may potentially serve as a proxy to predict the challenging nature of a topic before conducting any annotations for the topic, it should be considered carefully as there can indeed be other factors beyond similarity affecting the performance for a topic. Improving predictability of the difficulty of a topic could be an important avenue for future research.

\section*{Conclusions}
\label{sec:Conclusions}

Our research on check-worthy claim detection highlights the importance of considering a scenario which has been overlooked so far, i.e. tackling new topics which have not been seen before. Through the use of a competitive model, AraCW, we have shown that indeed tackling new topics is especially challenging. Given that this constitutes a realistic scenario where one has certain datasets labelled to train models and needs to deal with new topics for which to identify claims, our research calls for the need of more research in this scenario.

As a first attempt to tackle the cross-topic check-worthy claim detection task, we have proposed a novel model called AraCWA, which incorporates capabilities for few-shot learning and data augmentation into the claim detection. This model has led to substantial improvements which can reach up to 11\% overall improvements across multiple topics, positing this as a plausible direction to take for improving the cross-topic detection model.

Our research also opens important avenues for future work. Available datasets for claim detection are limited, not least in Arabic, which hinder broader investigation into wider sets of topics; expanding the sets of available topics would be highly valuable to enrich this research. Likewise, this could enable broader and deeper investigation into the inconsistent improvements across different topics, which could lead to further understanding into what makes a topic more or less challenging.


\bibliography{bibliography}

\appendix
\section{Appendix}
\begin{table}[ht]
\centering
\begin{tabular}{c|l|l}
SN &TopicID & Title\\\hline
1 & CT20-AR-05 & Protests in Lebanon \\
2 & CT20-AR-10 & Waseem Youssef \\
3 & CT20-AR-19 & Turkey's intervention   in Syria \\
4 & CT20-AR-01 & Deal of the century \\
5 & CT20-AR-02 & Houthis in Yemen \\
6 & CT20-AR-08 & Feminists  \\
7 & CT20-AR-12 & Sudan and   normalization \\
8 & CT20-AR-14 & Events in Libya \\
9 & CT20-AR-23 & The case of the Bidoon   in Kuwait \\
10 & CT20-AR-27 &  Algeria \\
11 & COVID-19 & COVID-19 \\
12 & CT20-AR-30 & Boycotting countries and spreading rumors against Qatar\\
13 & CT21-AR-01 & Events in Gulf \\
14 & CT21-AR-02 & Events in USA 
\end{tabular}
\caption{\label{tab:alltopics} Topics in CT20-AR and CT21-AR}
\end{table}

Note: The COVID-19 is obtained after combining the following topics from the original CT20-AR and CT21-AR datasets:

\begin{itemize}
 \item CT20-AR-03: New coronavirus
 \item CT20-AR-28\_w1: New Corona
 \item CT20-AR-28\_w2: New Corona
 \item CT20-AR-29: Corona in the Arab world
\end{itemize}

\end{document}